\documentclass[runningheads]{llncs}
\usepackage{graphicx}
\usepackage{amsmath}
\usepackage{amssymb}
\usepackage{algorithm}
\usepackage{algorithmic}

\begin{document}
\title{Data-Driven Randomized Learning of Feedforward Neural Networks\thanks{Supported by Grant 2017/27/B/ST6/01804 from the National Science Centre, Poland.}}
\author{Grzegorz Dudek\inst{1}\orcidID{0000-0002-2285-0327}}
\authorrunning{G. Dudek}
\institute{Electrical Engineering Department, Częstochowa University of Technology, Częstochowa, Poland\\	
	\email{dudek@el.pcz.czest.pl}}

\maketitle              % typeset the header of the contribution
\begin{abstract}
Randomized methods of neural network learning suffer from a problem with the generation of random parameters as they are difficult to set optimally to obtain a good projection space. The standard method draws the parameters from a fixed interval which is independent of the data scope and activation function type. This does not lead to good results in the approximation of the strongly nonlinear functions. In this work, a method which adjusts the random parameters, representing the slopes and positions of the sigmoids, to the target function features is proposed. The method randomly selects the input space regions, places the sigmoids in these regions and then adjusts the sigmoid slopes to the local fluctuations of the target function. This brings very good results in the approximation of the complex target functions when compared to the standard fixed interval method and other methods recently proposed in the literature.

%This is because the proposed method adjusts the slope of the sigmoids to the %local slopes of the target function and is able to control the bias-variance %tradeoff by its hyperparameters. 

\keywords{Data-driven randomized learning \and Feedforward neural networks \and Neural networks with random hidden nodes \and Randomized learning algorithms.}
\end{abstract}
\section{Introduction}
The learning of feedforward neural networks (FNNs) is an optimization process where the error function is highly nonconvex. Flat regions of the error function as well as many local minima and saddle points hinder and slow down the learning. This is because the gradient algorithms commonly used for FNN learning are very sensitive to the surface of the objective function and fall into the traps of the local minima. Moreover, the gradient calculations are time consuming, especially for a complex target function (TF), a big training data set or for a network with many hidden neurons and many layers. In randomized learning the gradient descend methods do not have to be used. The learning process is split into two stages. In the first stage, the parameters of the hidden neurons (single-hidden-layer FNN is considered), i.e. the weights and biases, are randomly selected. They are not trained at all and stay fixed. In the second stage, the output weights, connecting the hidden layer with the output layer, are trained. The optimization problem becomes convex and can be solved using a standard linear least-squares \cite{Pri15}, which is the simplest, most studied and scalable learning procedure to date.

The standard method of generating the parameters of the hidden neurons, i.e. the weights and biases, is to select them randomly with a fixed interval. It was theoretically proven that when the random parameters are selected from a symmetric interval according to any continuous sampling distribution, the FNN is a universal approximator \cite{Hau99}. The problem with selection of the appropriate interval for the parameters has not yet been solved, and is considered to be one of the most important research gaps in the field of randomized algorithms for NN training \cite{Zha16s}, \cite{Cao18}. In many practical applications of FNN with random parameters this interval is set as $[-1,1]$ without any justification, regardless of the problem type (classification or regression), data, and activation function type. Some works have shown that such an interval is misleading because the network is unable to model nonlinear maps, no matter how many training samples are provided or what size networks are used \cite{Li17}. So, the optimization of this interval is recommended for a specified task \cite{Pao94}, \cite{Hau99}. Such a way of improving the performance of the FNN with random parameters has been used in many works, e.g. \cite{Li19}.

In \cite{dud19} it was noted that the weights and biases have different meanings, i.e. weights represent the sigmoid slope and biases represent its shift, and therefore they should not be generated from the same interval. The  method proposed in \cite{dud19} generates the parameters of the hidden nodes in such a way that nonlinear fragments of the activation functions are located in the input space regions with data and can be used to construct a surface approximating a nonlinear TF. The weights and biases are dependent on the input data range and activation function type. This leads to an improvement in the approximation performance of the network. Another approach for generating the random parameters was proposed in \cite{dud19a}. This method, firstly, selects at random the slope angles of the sigmoids from the interval adjusted to the TF fluctuations, then rotates the sigmoids randomly and finally shifts them into the input space according to the data distribution. This gives much better results than the standard approach with fixed intervals.

In this work we do not select the hidden neurons parameters from specified intervals. Instead, we propose to adjust them to the local features of the TF. The proposed method selects the input space region by randomly choosing one of the training points, then places the sigmoid in this region and adjusts the sigmoid slope to the TF slope in the neighborhood of the chosen point. Combining linearly the randomly placed sigmoids in the input space, we obtain a fitted surface which reflects the TF features in different regions.    

\section{Randomized Learning of FNNs}
A single-hidden-layer FNN for a single output case and an input $\mathbf{x}=[x_1, x_2,...,\\ x_n]^T \subset \mathbb{R}^n$ is defined by linearly combining $m$ nonlinear transformations of the input $h_i(\mathbf{x})$:

\begin{equation}
\varphi(\mathbf{x}) = \sum_{i=1}^{m}\beta_ih_i(\mathbf{x}) = \mathbf{h}(\mathbf{x})\boldsymbol{\beta}
\label{eq1}
\end{equation}
where $\beta_i$ is the weight between the $i$-th hidden neuron and the output neuron, and $h_i(\mathbf{x})$ is represented by an activation function of the $i$-th hidden neuron, e.g. a sigmoid:

\begin{equation}
h_i(\mathbf{x}) = \frac{1}{1 + \exp\left(-\left(\mathbf{a}_i^T\mathbf{x} + b_i\right)\right)}
\label{eq2}
\end{equation}

In randomized FNNs, the weights $ \mathbf{a}_i = \left[ a_{i,1}, a_{i,2}, \ldots, a_{i,n}\right]^T $ and bias $ b_i$ are generated randomly for each neuron according to any continuous sampling distribution. Usually $ a_{i,j} \sim U(a_{min}, a_{max}) $ and $ b_i \sim U(b_{min}, b_{max} $).

Note that the activation function \eqref{eq2} applies some nonlinearity on a random linear combination of the input vector. As a result, the activation function is randomly located in the space, has random slope and rotation. 

The hidden layer output matrix for $N$ training samples is:

\begin{equation}\label{key}
\mathbf{H} = \left[
\begin{array}{c}
\mathbf{h}(\mathbf{x}_1) \\
\vdots \\
\mathbf{h}(\mathbf{x}_N) 
\end{array}
\right] =
\left[
\begin{array}{ccc}
h_1(\mathbf{x}_1) & \ldots & h_m(\mathbf{x}_1) \\
\vdots & \ddots & \vdots \\
h_1(\mathbf{x}_N) & \ldots & h_m(\mathbf{x}_N)
\end{array}
\right]
\end{equation}
where the $i$-th column of $\mathbf{H}$ is the $i$-th hidden node output vector with respect to inputs $ \mathbf{x}_1, \mathbf{x}_2, \ldots, \mathbf{x}_N$, and $ \mathbf{h}(\mathbf{x}) = \left[h_1(\mathbf{x}), h_2(\mathbf{x}), \ldots, h_m(\mathbf{x})\right] $ is a~nonlinear mapping from $n$-dimensional input space to $m$-dimensional feature space, wherein, typically $m\gg n$. 

The parameters of the hidden neurons, $a_{i,j}$ and $b_i$ are fixed, so, the matrix $\mathbf{H}$ is calculated only once and remains unchanged.

The output weights $\beta_i$ are determined by solving the following linear problem:
 
\begin{equation}
\mathbf{H}\boldsymbol{\beta} = \mathbf{Y}
\label{eq4}
\end{equation}
where $ \boldsymbol{\beta} = [\beta_1, \beta_2, \ldots, \beta_m]^T $ is a~vector of output weights and $ \mathbf{Y} = [y_1, y_2, \ldots, y_N]^T $ is a~vector of target outputs.

A least mean squares solution of \eqref{eq4} can be achieved within a single learning step by using the Moore–Penrose pseudoinverse $ \mathbf{H}^+ $ of matrix $ \mathbf{H} $: 

\begin{equation}
\boldsymbol{\beta} = \mathbf{H}^+\mathbf{Y}
\label{eq5}
\end{equation}

In the above described randomized learning there are five hyperparameters which influence strongly on the approximation abilities of the network. They are: the number of hidden nodes $m$ and the bounds of the scopes for weights and biases, i.e. $a_{min}, a_{max}, b_{min}$ and $b_{max}$. In most of the works on randomized algorithms for FNNs, the bounds of parameters are selected as fixed, regardless of the data and activation function types. Typically $ a_{min} = b_{min} = -1 $ and $ a_{max} = b_{max} = 1 $. The hidden neuron sigmoids, whose linear combination \eqref{eq2} builds the function fitting data, should deliver the nonlinear fragments, avoiding their saturated fragments, and so achieve the required accuracy of approximation. As demonstrated in \cite{dud19} when using typical interval for random parameters, $[-1, 1]$, the sigmoids are not distributed properly in the input space and their steepness does not correspond to the TF steepness. In some works, the authors optimize the interval for the random parameters by searching for its bounds $[-u,u]$.

To improve the randomized learning performance, the method proposed in \cite{dud19} randomly generates the weights and biases of the hidden nodes, depending on the input data range and activation function type, in such a way so as to introduce the nonlinear fragments of the activation functions in the input space region containing the data points. Additionally the slopes of the activation functions are adjusted to the TF complexity. According to the proposed method the weights of the $i$-th hidden node are calculated as follows:

\begin{equation}
a_{i,j} = \zeta_j\frac{\Sigma_i}{\sum\limits_{l=1}^n \zeta_l},\quad j=1,2,...,n
\label{eq6}
\end{equation}
where  $ \zeta_1, \zeta_2, \ldots, \zeta_n \sim U(-1, 1) $ are i.i.d. numbers and $\Sigma_i$ is the sum of weights of the $i$-th node, which is randomly chosen from the interval:   
\begin{equation}
\left|\Sigma_i\right| \in \left[ \ln\left( \frac{1-r}{r}\right), s\cdot\ln\left( \frac{1-r}{r}\right) \right]
\label{eq7}
\end{equation}

There are two parameters which decide on the activation function slope: $r\in(0,0.5)$ and $s>1$. Specifically, they determine two boundary sigmoids between which the activation functions are randomly generated. These parameters are adjusted to the data in cross-validation. 
   
The biases of the hidden nodes are determined setting the inflection points of the sigmoids at some points $\mathbf{x}^*$ randomly selected from the input space or, alternatively, randomly chosen from the training set. The bias of the $i$-th node is calculated as follows:

\begin{equation}
b_i = -\mathbf{a}_i^T\mathbf{x}^*
\label{eq8}
\end{equation} 

From the above equations we can conclude that the new approach to selection of the random parameters is a radical departure from the standard approach. Instead of generating both weights and biases from the fixed interval, in the new approach, we first generate the sum of the all node weights from the interval \eqref{eq7}, and then randomly generate individual weights from \eqref{eq6}. In the next step, the bias is generated from \eqref{eq8} on the basis of randomly chosen point $\mathbf{x}^*$ and weight vector $\mathbf{a}_i$. The derivations of the above equations and more detailed discussion on this topic, including other activation function types, can be found in \cite{dud19}.

Another method for improving the performance of FNN randomized learning was proposed in \cite{dud19a}. Firstly, it randomly choses the slope angles of the hidden neurons activation functions from an interval adjusted to the complexity of the TF. Then, the activation functions are randomly rotated around the y-axis and finally, they are distributed across the input space according to data distribution. For complex TFs, with strong fluctuations, the proposed
method gives incomparably better results than the standard approach with the fixed interval for the random parameters. This is because it adjusts the slopes of the activation functions to the data and introduces their steepest fragments into the input space, avoiding their saturation fragments. 

In this approach, the weights of the $i$-th hidden node are calculated from:

\begin{equation} 
a_{i,j} = -4\frac{a_{i,j}^{\prime}}{a_{i,0}^{\prime}},\quad j=1,2,...,n
\label{eq9}
\end{equation} 
where $a_{i,j}^{\prime}$ are components of the normal vector $\mathbf{n}$ to the hyperplane, which is tangent to the sigmoid at their inflection points.

The angle between the normal vector $\mathbf{n}$ and the unit vector in the direction of the y-axis, $\alpha$, is randomly selected from the interval $(\alpha_{min}, \alpha_{max})$. The bounds of this interval are adjusted to the TF in cross-validation. These bounds control the slopes of the sigmoids and thus the flexibility of the model. The rotation of the individual sigmoid is random, determined by choosing randomly the components of the normal vector $a_1^{\prime}, \ldots, a_n^{\prime} \sim U(-1, 1)$ and calculating component $a_0^{\prime}$ from:       

\begin{equation}
a_0^{\prime} = (-1)^c \frac{\sqrt{(a_1^{\prime})^2+...+(a_n^{\prime})^2}}{\tan \alpha}
\end{equation}
where $c \sim U\{0, 1\}$.

To distribute the sigmoids across the input space their biases are calculated from \eqref{eq8}, where  
$\mathbf{x}^*$ are the randomly chosen training points. Details of this method can be found in \cite{dud19a}.

\begin{figure}
 	\centering
	\includegraphics[width=0.49\textwidth]{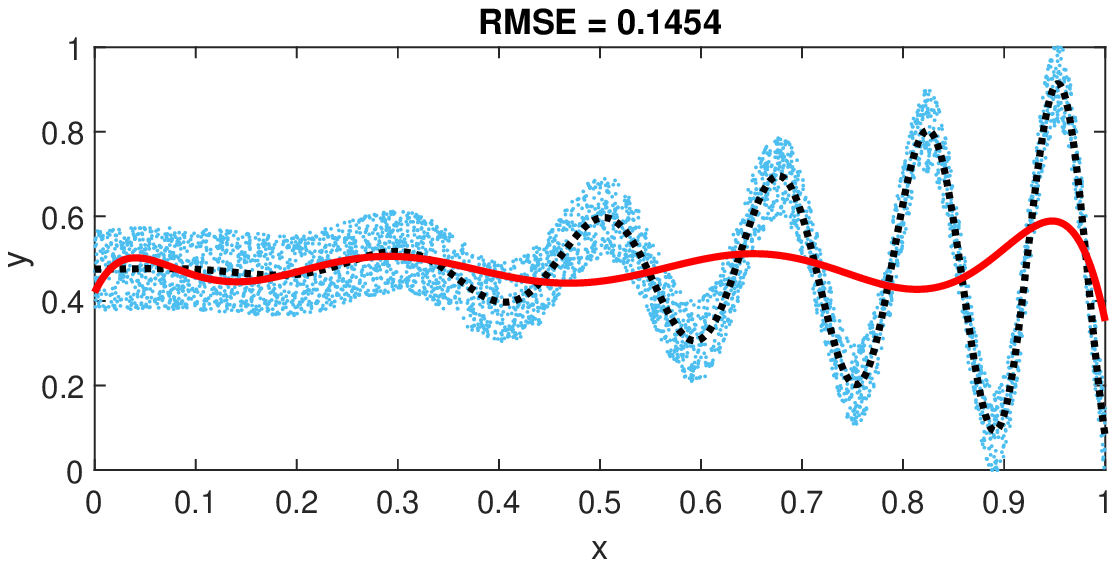}
	\includegraphics[width=0.49\textwidth]{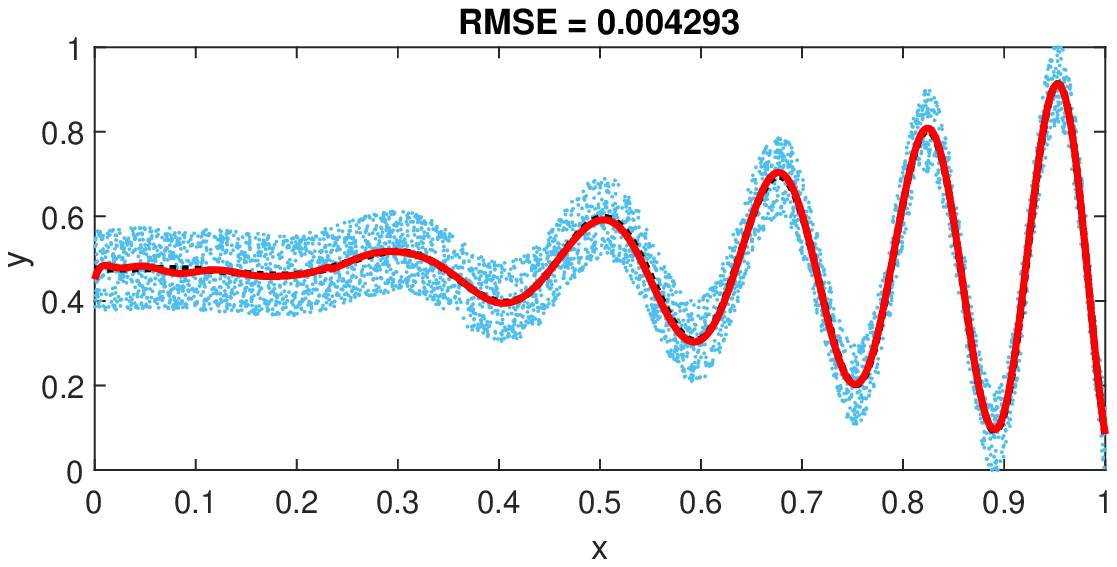}
	\includegraphics[width=0.49\textwidth]{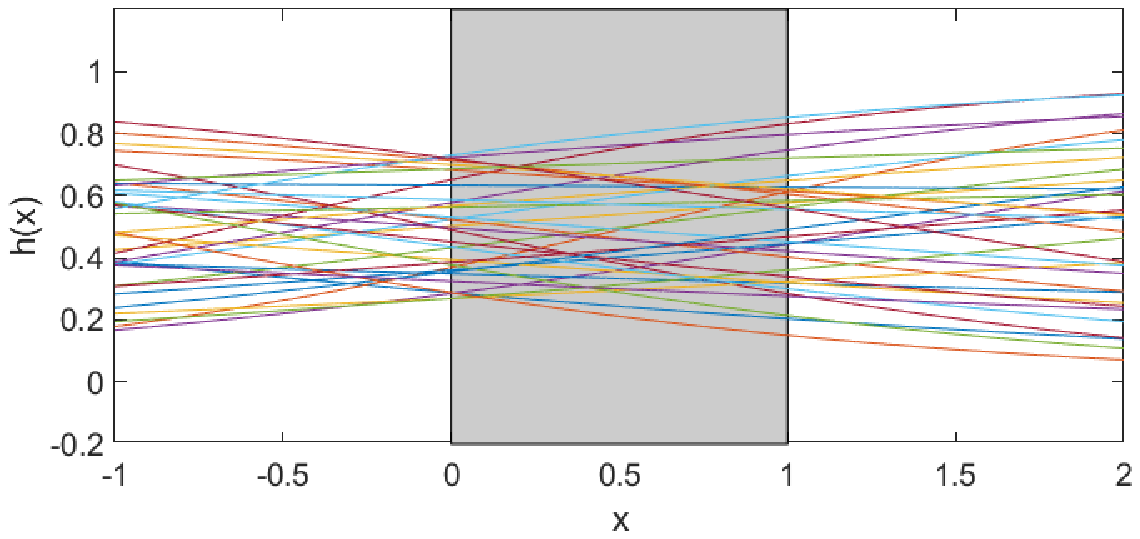}
	\includegraphics[width=0.49\textwidth]{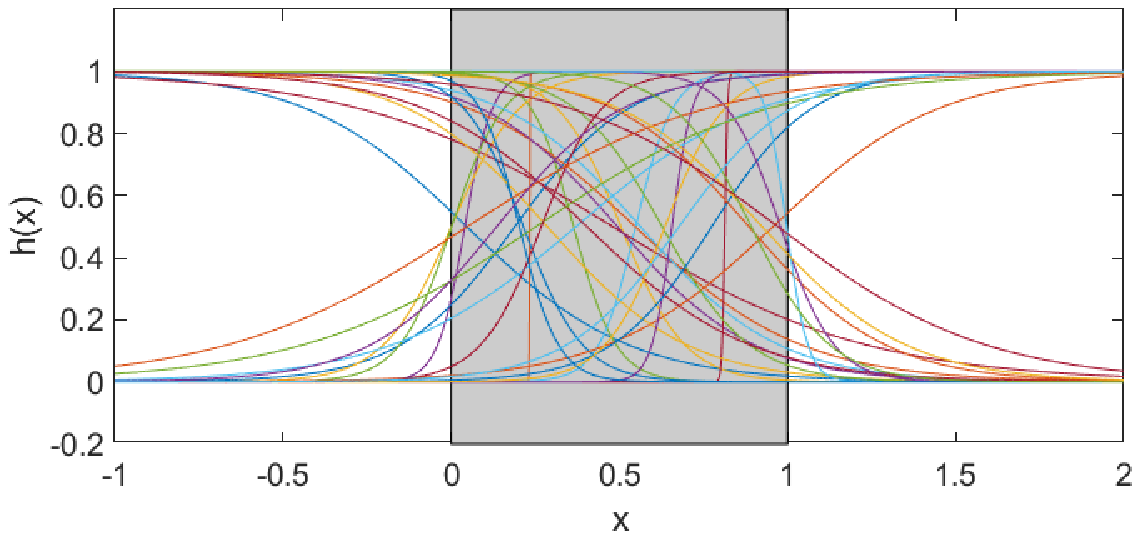}
	\includegraphics[width=0.49\textwidth]{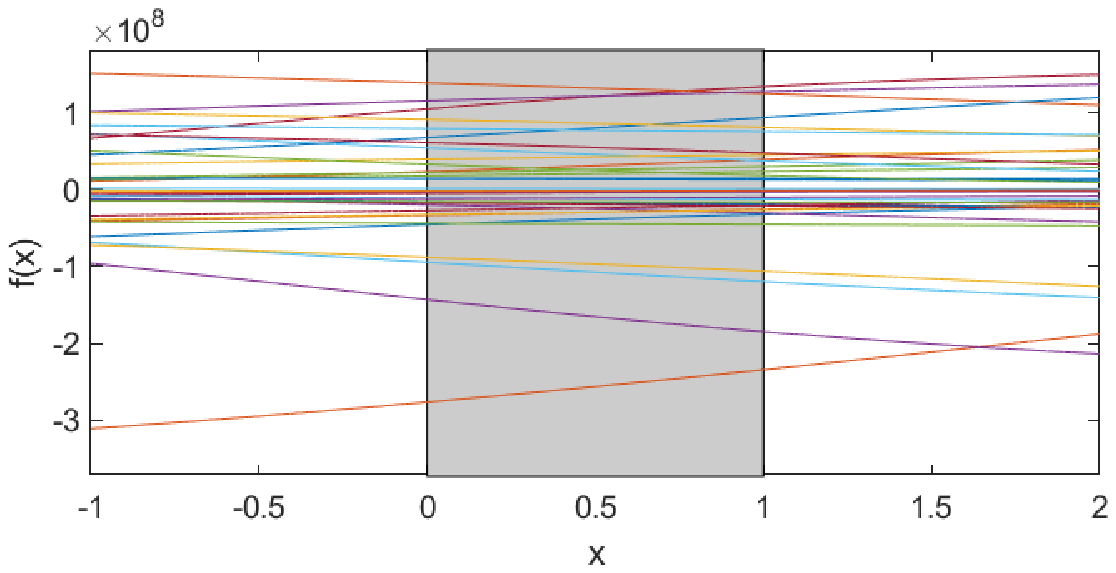}
	\includegraphics[width=0.49\textwidth]{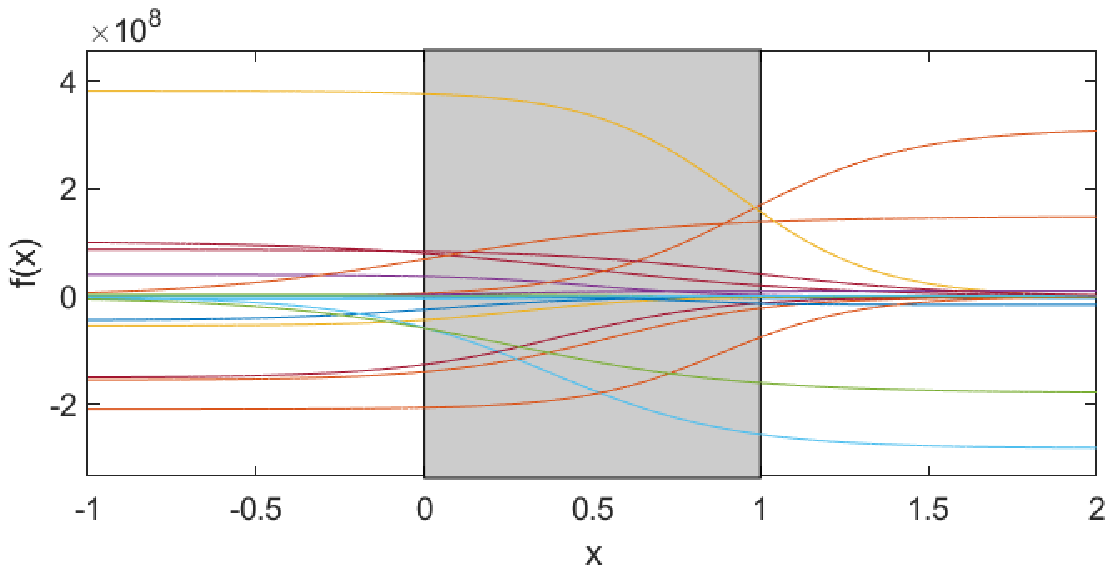}
	\caption{Fitted curves (upper panel), hidden node sigmoids (middle panel) and weighted sigmoids
		(bottom panel) for the standard method (left panel) and the RARSM method (right panel).} \label{fig1}
\end{figure}

The methods proposed in \cite{dud19} and \cite{dud19a} allow us to control the slope of the sigmoids forming the fitted function, and hence the degree of generalization of the network and bias-variance tradeoff of the model. In Fig. \ref{fig1}, the approximation of the highly nonlinear function is shown for the standard method of generating random parameters from the fixed intervals $[-1,1]$ and for the method proposed in \cite{dud19a} (let us denote this method with the acronym RARSM, i.e. random sigmoid slope angle, rotation and shift method). The TF is in the form: 

\begin{equation}
g(x) = \sin\left(20\cdot\exp(x)\right)\cdot x^2
\label{eqTF1}
\end{equation}

The FNNs learns from a training set containing $ 5000 $ points $ (x_l, y_l) $, where $ x_l $ are uniformly randomly distributed on $ [0, 1] $ and $ y_l $  are calculated from \eqref{eqTF1} and then distorted by adding the uniform noise distributed in $ [-0.2, 0.2] $. The test set of the same size expresses the true TF \eqref{eqTF1}. The outputs are normalized in the range $ [0, 1] $.

In both cases $35$ hidden neurons were used. In RARSM $\alpha_{min}=30^\circ$ and $\alpha_{max}=90^\circ$. The sigmoids distributed in the input interval, which is shown as a gray field, are shown in the middle panel of Fig. \ref{fig1}. After weighing them by the output weights $\beta_i$ we obtain the curves shown in the bottom panel. The sum of these curves gives the fitted functions, which are drawn with solid lines in the upper panel. Note that the sigmoids generated from the fixed internal $[-1,1]$ are too flat and their steepest fragments, around their inflection points, do not correspond to the steep fragments of the TF. As a result, they cannot be combined to obtain the TF, even when we increase the number of neurons to several hundreds or even thousands. In a completely different manner the sigmoids are generated by the RARSM. As we can see from the middle right panel of Fig. \ref{fig1}, the sigmoids have their steepest fragments inside the input interval. Their slopes are also fitted to the TF. This results in a reduction the error from $RMSE=0.1454$ for the fixed interval to $RMSE=0.0043$.       

\section{Data-Driven Generation of Hidden Nodes Parameters}
In the above described methods of FNN randomized learning, we can observe an evolution. The first step of this evolution is the completely random generation of the hidden node weights and biases, both from a fixed interval, typically $[-1, 1]$, according to any continuous sampling distribution, usually a uniform one. In the second step, we try to find the best interval for the weights so as to match the slopes of sigmoids to the TF complexity. After adjusting slopes, we calculate the biases in order to distribute the sigmoids randomly across the input space. The interval for the weights is dependent on the two hyperparameters, $r$ and $s$, which are searched in the cross-validation. In the third step of the evolution, we introduce hyperparameters describing the sigmoid shape which are more intuitive than $r$ and $s$, i.e the limit slope angles of the sigmoids. These hyperparameters can be adjusted to the TF complexity or their default values can be applied, $\alpha_{min}=0^\circ$ and $\alpha_{max}=90^\circ$. The forth step of the evolution, proposed in this work, is adjusting the sigmoids individually to the local complexity of the TF. The weights and biases of the sigmoids are no longer random, but they are adjusted to data in randomly chosen regions of the input space. So, each sigmoid models locally the TF in the neighborhood of a randomly selected training point. The fitted function is constructed typically as a linear combination of the sigmoids \eqref{eq1}. The weights in this combination are calculated according to \eqref{eq8} using the Moore–Penrose pseudoinverse.
The idea behind the proposed method is shown on the single-variable TF example in subsection 3.1, and its multivariable extension is presented in subsection 3.2.  

\subsection{The Idea behind the Method}
The idea behind the proposed method can be exemplified by the approximation of the single-variable TF \eqref{eqTF1}. Sigmoids are used as hidden nodes activation functions:

\begin{equation}
h(x) = \frac{1}{1 + \exp(-(ax + b))}
\label{eqSig}
\end{equation}
where $ a $ is a weight controlling a~slope of the sigmoid and $ b $ is a bias shifting the sigmoid along the x-axis. 

It would be convenient to place the sigmoids in the input space and adjust their slopes in such a way that they correspond to the TF fluctuations. To do so, let us select randomly training point $x^*$ and find its $k$ nearest neighbors. These $k+1$ points form the neighborhood of $x^*$, $\Psi(x^*)$, and express the local features of the TF around $x^*$.  Now, let us fit to these points straight line $T$:
    
\begin{equation}
y = a'x + b'
\label{eqLin}
\end{equation}
Note that coefficient $a'$ expresses the slope of the line which corresponds to the slope of the TF in $\Psi(x^*)$. 

Let set some sigmoid $S$ in the input space in such a way that its inflection point $P$ is in $x^*$. Remembering that the sigmoid value for the inflection point is $0.5$, we get: 

\begin{equation}
h(x^*) = \frac{1}{1 + \exp(-(ax^* + b))}=0.5 
\label{eqSig2}
\end{equation}
Let us assume that the slope of $S$ at $P$ is the same as the slope of the line $T$. This means that the derivative of $S$ at $P=x^*$ is equal to the derivative of the line $T$, thus:    

\begin{equation}
ah(x^*)(1-h(x^*))=a' 
\label{eqDer1}
\end{equation}
Substituting $h(x^*) = 0.5$ from \eqref{eqSig2} into \eqref{eqDer1} we obtain:

\begin{equation}
a\cdot0.5\cdot(1-0.5) = a'
\label{eqDer3}
\end{equation}
and finally weight $a$ of sigmoid $S$ is:
 
 \begin{equation}
 a = 4a'
 \label{eqDer4}
 \end{equation}
 
 From \eqref{eqSig2} we also obtain:
 
\begin{equation}
b = -ax^*
 \label{eqDer5}
\end{equation}

So, the sigmoid which models locally the TF in the neighborhood of $x^*$ has weight $a$ dependent on slope parameter $a'$ of the line fitted to $\Psi(x^*)$. Additionally bias $b$ is dependent on slope parameter $a'$ as well as on point $x^*$. 
    
\begin{figure}
 	\centering
 	\includegraphics[width=0.6\textwidth]{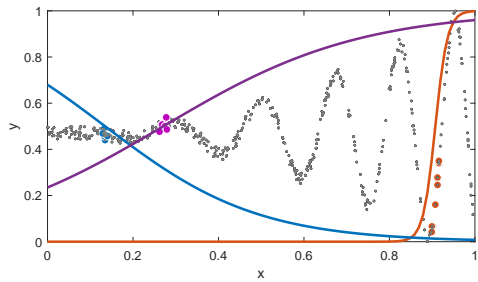}
 	\caption{Examples of setting the sigmoids to the neighborhoods (larger colored points) of the three randomly selected training points.} \label{fig2}
 \end{figure}

Fig.~\ref{fig2} shows an example sigmoids set according to the three randomly selected points $x^*$ and their neighborhoods composed of $x^*$ and their ten closest points. Note that the sigmoids reflect the slopes of the TF around points $x^*$.      
 
\begin{figure}
 	\centering
	\includegraphics[width=0.49\textwidth]{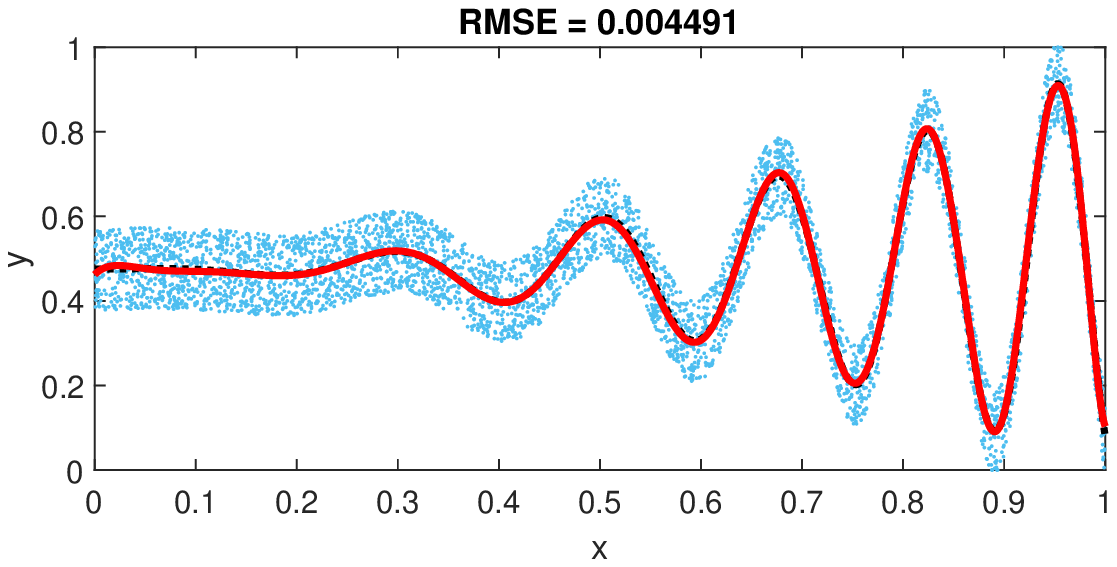}\\
	\includegraphics[width=0.49\textwidth]{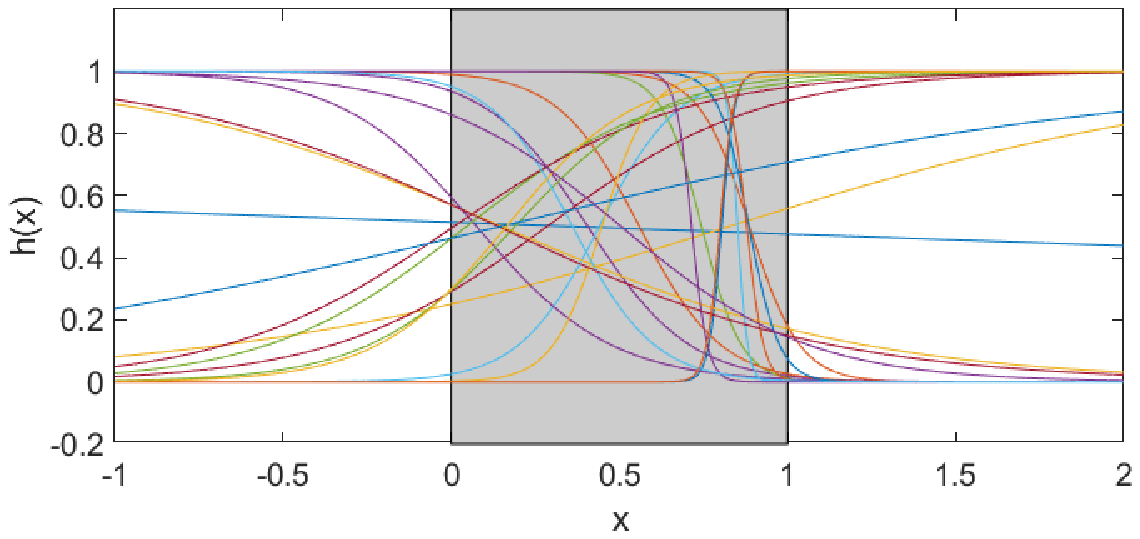}\\
	\includegraphics[width=0.49\textwidth]{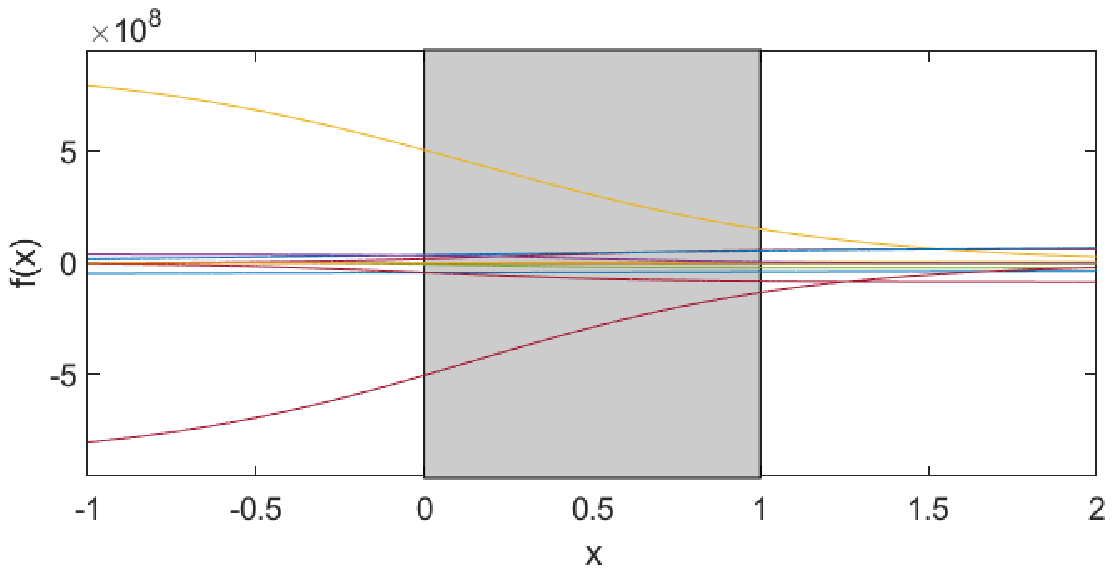}
	\caption{Fitted curve (upper panel), hidden node sigmoids (middle panel) and weighted sigmoids
		(bottom panel) for the proposed method.} 
	\label{fig3}
\end{figure}

Fig.~\ref{fig3} shows function \eqref{eqSig} approximation when using the proposed method with 25 hidden neurons and 100 nearest neighbors. Compare Fig.~\ref{fig3} with Fig.~\ref{fig1} and note the similar errors for the proposed method and RARSM. Note also the different distribution of the the sigmoids which have different slopes in these both cases. In the proposed method the steeper sigmoids are generated at the right border of the input interval, where the fluctuations of the TF are stronger. At the left border, where the TF is flat, the sigmoids are less steep. While in the RARSM the steepness of the sigmoids does not depend on the local TF features and is similar in each region of the input space. 

It is worth mentioning that the proposed method needs only 25 hidden neurons to obtain the same error level as the RARSM with 35 neurons. The RMSE for the RARSM with 25 neurons was $0.031$, which is almost seven times larger than for the proposed method. The smaller number of neurons in the proposed approach is due to the fact that the sigmoids are better fitted to TF fluctuations.   

%This can translate into unwanted fluctuations of the FF on the flat TF fragments. This case is visualized in Fig.~\ref{fig4}, where we can observe smoother fitting for the proposed methods than for SA method. So, for TFs having variable fluctuations in the input space, the proposed method limits overfitting on the TF fragments with smaller fluctuations.
% \begin{figure}
%	\centering
%	\includegraphics[width=0.8\textwidth]{Fit1D_4.eps}
%	\caption{Comparison of the FF smoothness on a flat TF fragment for SA and the proposed methods.} %\label{fig4}
%\end{figure} 

%kąty 25 neur-0.03099, 100 - 0.00592
%std 25-0.1444, 35-0.1454, 50-0.1454, 100,500,1000 -0.1422

\subsection{Multivariable Function Fitting}
In this subsection, the proposed method is extended to the general case of multivariable function fitting. In this case the TF is a function of $n$ input variables included in the vector $\mathbf{x}=[x_1, x_2,..., x_n]^T \subset \mathbb{R}^n$. Similarly to the single-variable case, we place the sigmoids in the input space and adjust their slopes in such a way that they correspond to the TF fluctuations. The slopes of the TF in some point $\mathbf{x}^*$ are approximated by hyperplane $T$ fitted to the neighborhood $\Psi(\mathbf{x}^*)$ which includes this point and its $k$ nearest neighbors among the training points. This hyperplane is of the form:
     
\begin{equation}
y = a_1'x_1 + a_2'x_2+...+a_n'x_n +b'
\label{eqLin}
\end{equation}
where coefficient $a_j'$ expresses a slope of hyperplane $T$ in the $j$-th direction. 

Let us consider a sigmoid $S$ which has one of its inflection points, $P$, in the randomly selected training point $\mathbf{x}^*$:

\begin{equation}
h(\mathbf{x}^*) = \frac{1}{1 + \exp\left(-\left(\mathbf{a}^T\mathbf{x}^* + b\right)\right)}=0.5
\label{eqSigM}
\end{equation}
where $\mathbf{a}=[a_1, a_2,..., a_n]^T \subset \mathbb{R}^n$.

The slope of this sigmoid $S$ at point $\mathbf{x}^*$ in the $j$-th direction is expressed by a partial derivative:

\begin{equation} 
\frac{\partial h(\mathbf{x}^*)}{\partial x_j} = a_j h(\mathbf{x}^*)(1-h(\mathbf{x}^*))
\label{eqDerM}
\end{equation}

We want the slope of sigmoid $S$ at $\mathbf{x}^*$ in the $j$-th direction to be the same as the slope of the TF in this point, which is approximated by the hyperplane $T$ slope $a_j'$. Thus:

 \begin{equation} 
 a_j h(\mathbf{x}^*)(1-h(\mathbf{x}^*)) = a_j'
 \label{eqDerM1}
 \end{equation} 

After substituting $h(\mathbf{x}^*)=0.5$ from \eqref{eqSigM} into \eqref{eqDerM1} we obtain:

 \begin{equation}
a_j = 4a_j', \quad  j = 1, 2, ..., n
\label{eqDer4a}
\end{equation}

Directly from \eqref{eqSigM} we also obtain:

\begin{equation}
b = -\mathbf{a}^T\mathbf{x}^*
\label{eqDer5a}
\end{equation}

Note that the weights of the hidden neuron, $a_j$, expressing the slopes of the sigmoid in all $n$ directions, are proportional to the hyperplane $T$ coefficients corresponding to these directions. These coefficients approximate the TF slopes at the randomly selected point $\mathbf{x}^*$. The bias of sigmoid $S$ is a linear combination of point $\mathbf{x}^*$ and sigmoid weights $\mathbf{a}$. The sigmoid $S$ reflects the local features of the TF around point $\mathbf{x}^*$. Selecting randomly a set of points $\mathbf{x}^*$ we generate a set of sigmoids reflecting the local features of the TF in different regions. These sigmoids are the basis functions which are linearly combined to get the fitted function approximating the TF. The weights in this combination are calculated using the Moore–Penrose pseudoinverse \eqref{eq8}.

The proposed method places the sigmoids in the input space setting their inflection points on the randomly selected training points $\mathbf{x}^*$ and adjusting the sigmoid slopes to the slopes of the TF around these points. The TF slope is approximated by hyperplane $T$ fitted to the neighborhood of $\mathbf{x}^*$, i.e. $\mathbf{x}^*$ and its $k$ nearest neighbors. The TF is defined in $(n+1)$-dimensional space. To define hyperplane $T$ in such a space at least $n+1$ points are needed. So, the number of nearest neighbors $k$ should not be less than $n$. 

The number of nearest neighbors controls the bias-variance tradeoff of the model. The optimal value of $k$ depends on the random error observed in the data and the TF complexity. When the training points represent a TF with low error, the number of nearest neighbors $k$ should be lower. For higher errors a low value of $k$ leads to overfitting. On the other hand, a too large $k$ causes underfitting. This hyperparameter should be tuned in the cross-validation to the given data as well as the second hyperparameter, the number of hidden nodes $m$. In the experimental part of the work, the impact of the noise level in the data on the hyperparameters is investigated.

Algorithm 1 summarizes the proposed method.

\begin{algorithm}[H]
	\caption{Data-Driven Generating the Parameters of FNN Hidden Nodes}
	\label{alg1}
	\begin{algorithmic}
		\vspace{4mm}
		\STATE {\bfseries Input:}\\ 
		\vspace{4mm}
		\hspace{4mm} Number of hidden nodes $m$\\
		\hspace{4mm} Number of nearest neighbors $k\geq n$\\
		\hspace{4mm} Training set $ \Phi = \left\lbrace (\mathbf{x}_l, y_l) | \mathbf{x}_l \in \mathbb{R}^n, y_l \in \mathbb{R}, l = 1, 2,\ldots, N \right\rbrace $\\
	
		\vspace{4mm}
		\STATE {\bfseries Output:}\\ 
		\vspace{4mm}
		\hspace{4mm} Weights $ \mathbf{A} = \left[
		\begin{array}{ccc}
		a_{1,1} & \ldots & a_{m,1} \\
		\vdots & \ddots & \vdots \\
		a_{1,n} & \ldots & a_{m,n}
		\end{array}
		\right]	$  \\
		\hspace{4mm} Biases $ \mathbf{b} = [b_1, \ldots, b_m] $ \\    
		\vspace{4mm}
		\STATE {\bfseries Procedure:}\\
		\vspace{4mm}
		\FOR{$i=1$ {\bfseries to} $m$}
		\STATE Choose randomly $\mathbf{x}^* = \mathbf{x}_l \in \Phi $, 
		where $l \sim U\{1, 2, \ldots, N\}$ \\
		\STATE Create the set $\Psi(\mathbf{x}^*)$ containing $\mathbf{x}^*$ and its $k$ nearest neighbors in $\Phi $
		\STATE Fit the hyperplane to $\Psi(\mathbf{x}^*)$: 
		\begin{equation*}
		y = a_1'x_1 + a_2'x_2+...+a_n'x_n +b'
		\end{equation*}
		\STATE Calculate the weights for the $i$-th node:
		\begin{equation*}
		a_{i,j} = 4a_j', \quad  j = 1, 2, ..., n  
		\end{equation*}
		\STATE Calculate the bias for the $i$-th node:
		\begin{equation*}
		b_i = -\sum\limits_{j=1}^n a_{i,j}x_j^*
		\end{equation*}
		\ENDFOR
	\end{algorithmic}
\end{algorithm}

\section{Simulation study}
This section reports some experimental results of the proposed method, including the impact of the noise level in data on the hyperparameters and performance evaluation. 
In the first experiment we analyze how the noise disturbing data affects the hyperparameters of the proposed methods. Simulations were carried out on a two-variable TF of the form: 
\begin{equation}
g(\mathbf{x}) = \sin\left(20\cdot \exp\left(x_1\right)\right)\cdot x_1^2 + \sin\left(20\cdot \exp\left(x_2\right)\right)\cdot x_2^2
\label{eq25}
\end{equation}
On the basis of this function, training set $\Phi$ was created containing 5000 points $(\mathbf{x}_l, y_l)$, where the components of $ \mathbf{x}_l$ are independently uniformly randomly distributed on $[0, 1]$ and $y_l$ are generated from \eqref{eq25}, then normalized to the range $[0, 1]$ and finally distorted by adding the uniform noise distributed in $[–c, c]$. The testing set represents the TF without noise normalized into $[0, 1]$. It contains 10,000 points distributed regularly on a grid in the input space. 

\begin{figure}
	\centering
	\includegraphics[width=0.8\textwidth]{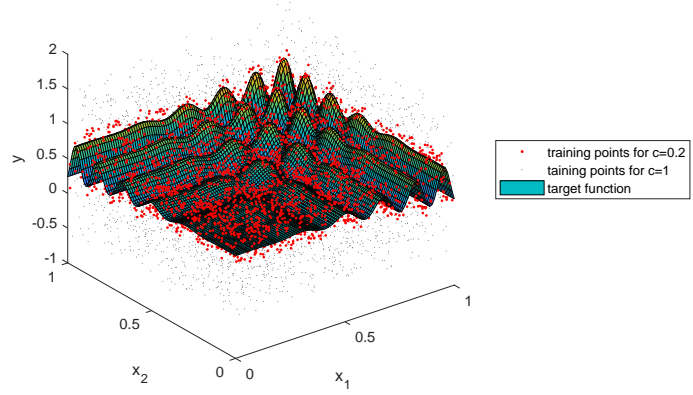}
	\caption{Target function and the training points for $c=0.2$ and $c=1$.} 
	\label{fig4}
\end{figure}

To introduce the noise of different level to data we changed the noise boundary $c$ from $0$ to $1$ with steps of $0.1$. This translates into a noise level from $0$ to $100\%$, defined as the ratio of the noise range to the TF range. The TF and the data points for two noise levels, $c=0.2$ and $c=1$, are shown in Fig. \ref{fig4}. For each noise level we changed the neighborhood $\Psi(x^*)$ size, $k'=3,5,7,10,20,...,100$, where $k'=k+1$, keeping the fixed number of hidden nodes $m=300$. For each setting, $100$ independent trials of FNN training were performed. 

\begin{figure}
	\centering
	\includegraphics[width=0.49\textwidth]{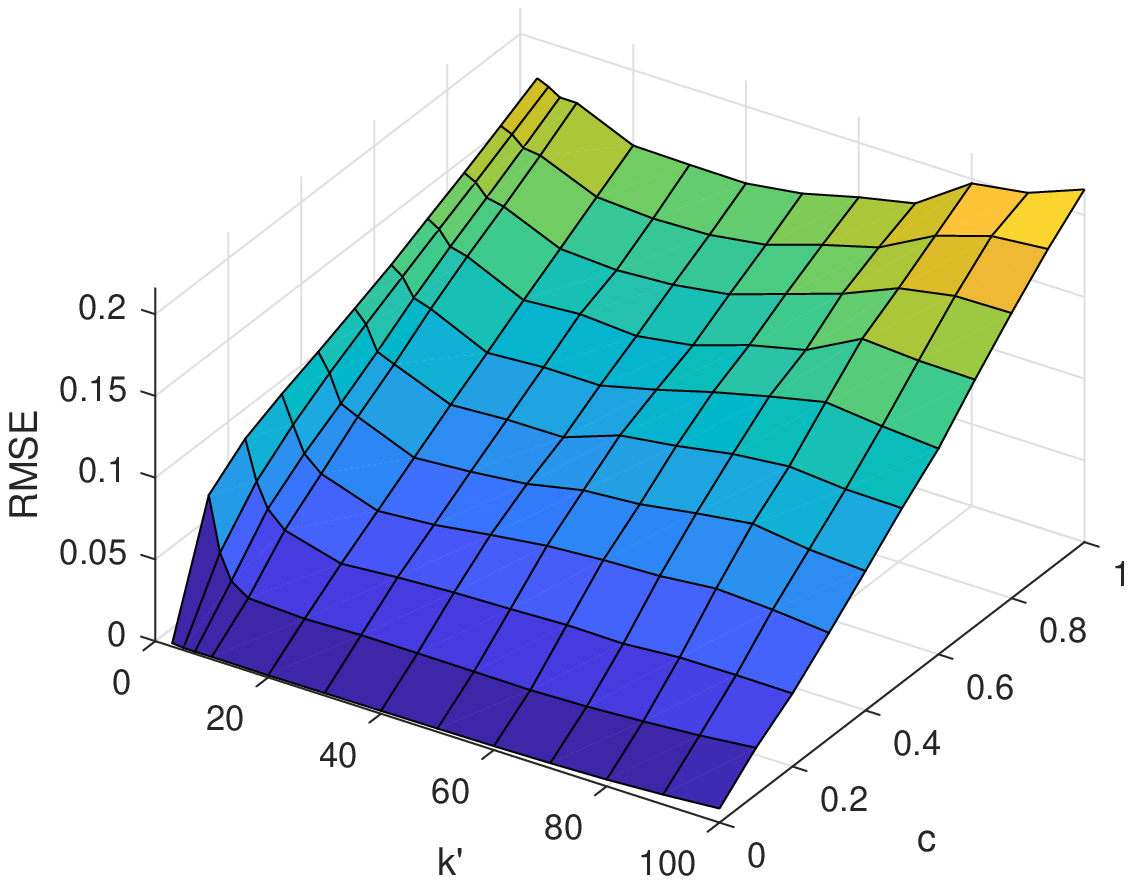}
	\includegraphics[width=0.49\textwidth]{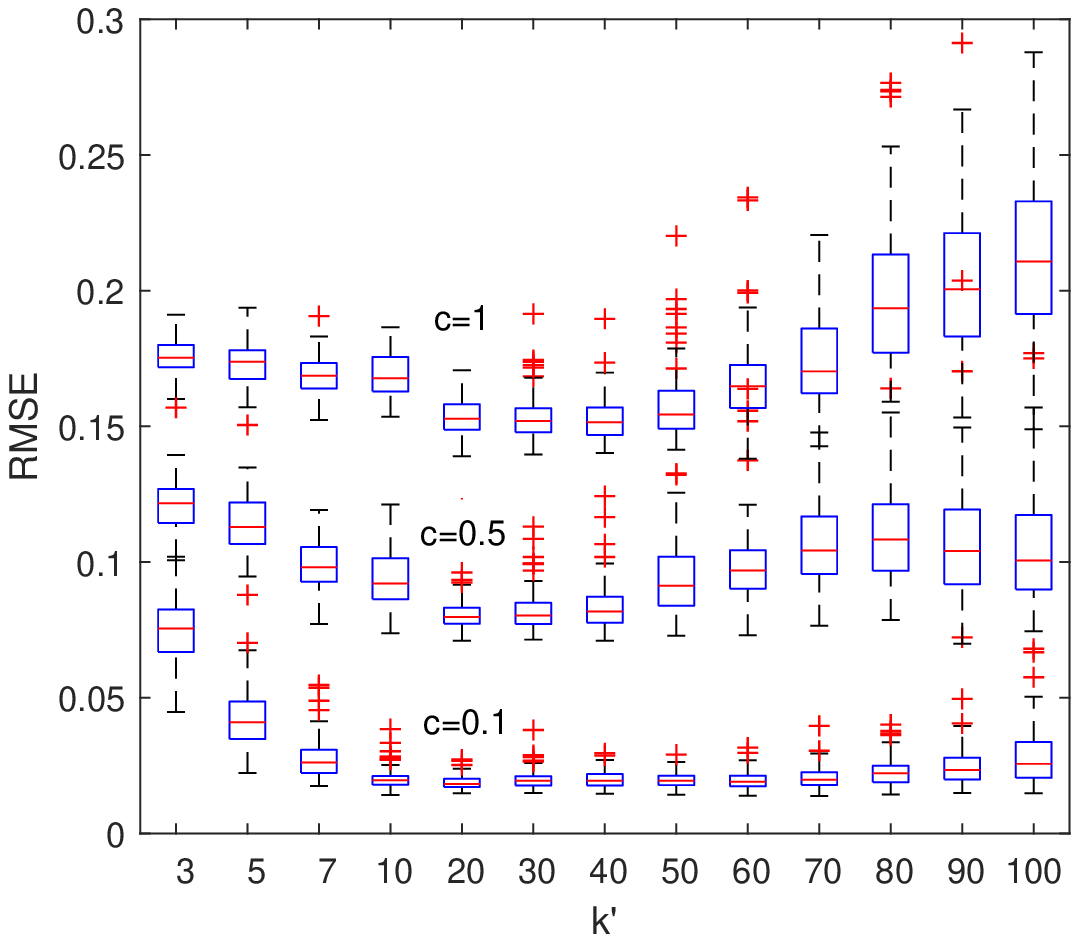}
	\caption{RMSE for different neighborhood size $k'$ and noise level $c$ at $m=300$.} 
	\label{fig5}
\end{figure}

The left panel of Fig. \ref{fig5} shows the test root-mean-square error (RMSE) for a different noise levels and neighborhood size. On the right panel, the boxplots are shown for three noise levels: $c=0.1, c=0.5$ and $c=1$. The optimal neighborhood size $k'$ was $20$ for the lower noise levels ($c\leqslant0.5$) and $30$ for the higher noise levels ($c>0.5$). The model tends to overfit for the lower than the optimal values of $k'$, and for higher values it tends to underfit. 

\begin{figure}
	\centering
	\includegraphics[width=0.49\textwidth]{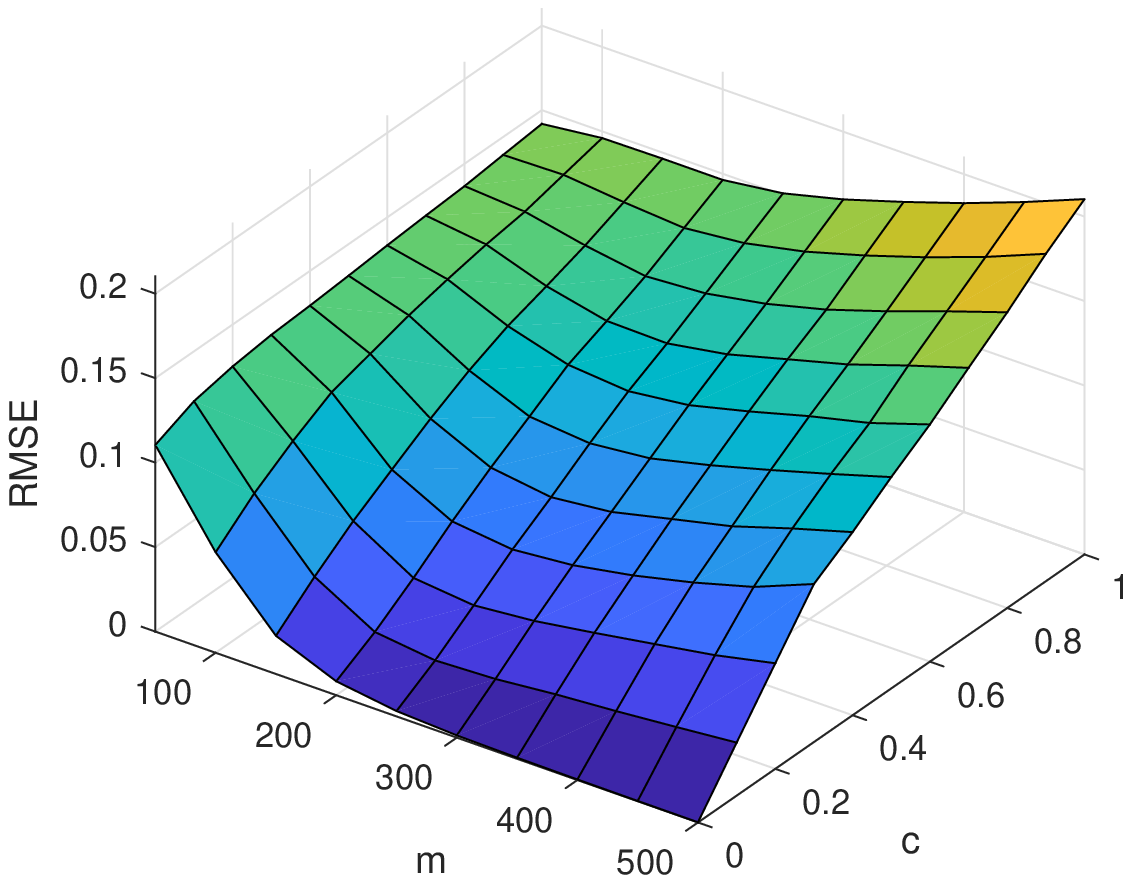}
	\includegraphics[width=0.49\textwidth]{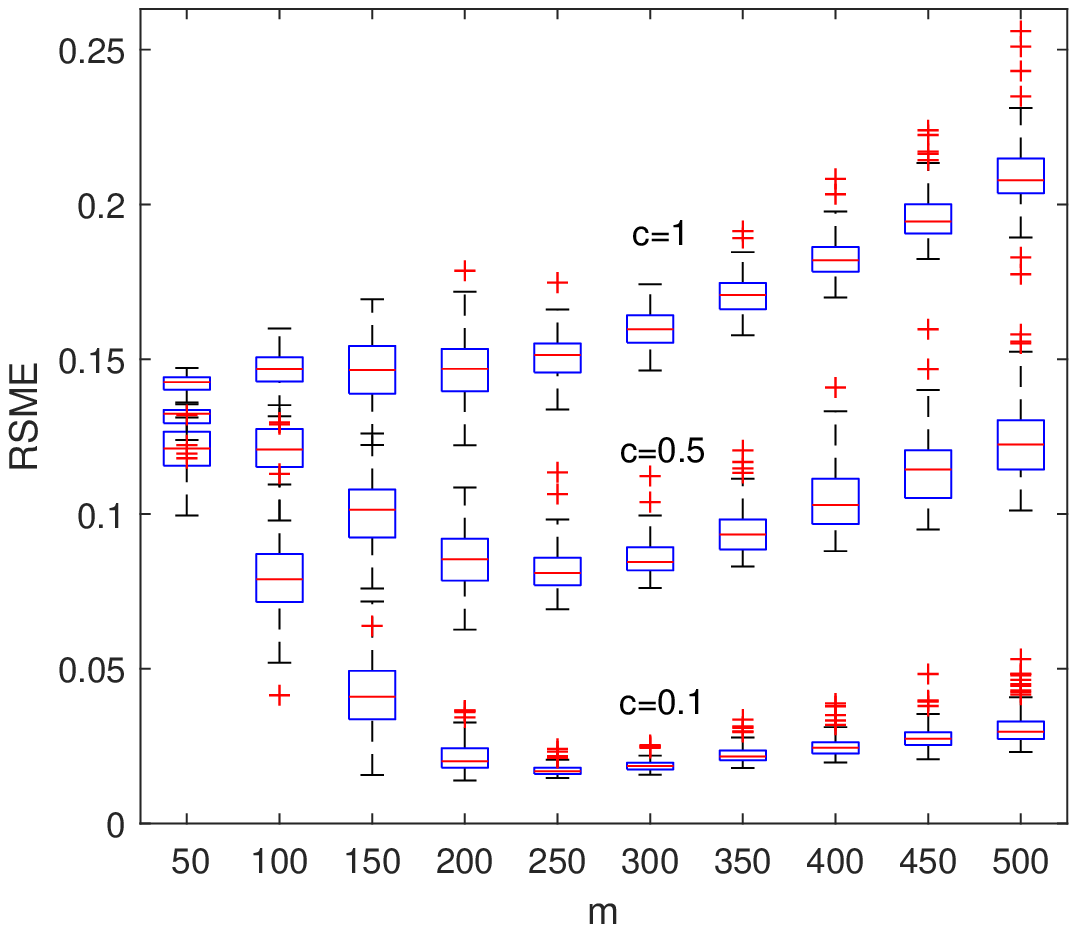}
	\caption{RMSE for different number of nodes $m$ and noise level $c$ at $k'=20$.} 
	\label{fig6}
\end{figure}

In the next step, for each noise level we changed the number of hidden nodes $m=50,100,...,500$ keeping a fixed size of the neighborhood $k'=20$. Fig. \ref{fig6} shows the test RMSE surface and the boxplots for this case. The optimal node numbers were: $m=250$ in the broad range of noise level from $0.1$ to $0.7$, $m=200$ for $c=0.8$ and $c=0.9$, and $m=50$ for $c=1$. We can observe from Figs. \ref{fig5} and \ref{fig6} that when the noise level is small, the underestimation of both hyperparameters, $k'$ and $m$, is more disadvantageous in terms of the error than overestimation.

%dla 1D pokazać wyniki dla k=1,... i m=1,..
%prudnik kopa biskupia
        
In the next experiments, we compare the results of the proposed method with the methods described in Section 2:

\begin{itemize}
	\item FIM - fixed interval method, standard method with fixed interval for the random parameters $[-1,1]$,
	\item OIM - optimized interval method, the method with the optimized interval for the random parameters $[-u,u]$, where $u$ in our simulations was selected from a given set $\{0.1, 0.5, 1, 2, \\
	3, 4, 5, 10, 15, 20, 30, 40, 50, 100, 200, 300, 400, 500\}$,
	\item rsM - the method proposed in \cite{dud19} with two parameters, $r$ and $s$, which were selected from the sets: $r\in\{0.0001, 0.001, 0.01, 0.015, 0.02,	0.3, 0.4, 0.5\}$ and $s\in\{2, 4, 6, 8, 10, 20, 30\}$,
	\item RARSM - random slope angle, rotation and shift method,  proposed in \cite{dud19a} with two parameters selected from the sets: $\alpha_{min}\in\{0^\circ, 5^\circ, ..., 85^\circ\}$ and  $\alpha_{max}\in\{\alpha_{min}+5^\circ,\alpha_{min}+10^\circ, ..., 90^\circ\}$,
	\item D-DM - data-driven method proposed in this work with parameter $k'$ which was selected from the set $\{5, 10, ..., 50\}$. 
\end{itemize}

For each method of random parameter generation, the number of hidden nodes was selected from the set $\{50,100, ..., 1000\}$. The selection of the optimal hyperparameter values was carried out in the grid search procedure using 10-fold cross-validation. For the optimal values of the hyperparameters $100$ independent trials of training were performed and test errors were calculated. 

\begin{figure}
	\centering
	\includegraphics[width=0.49\textwidth]{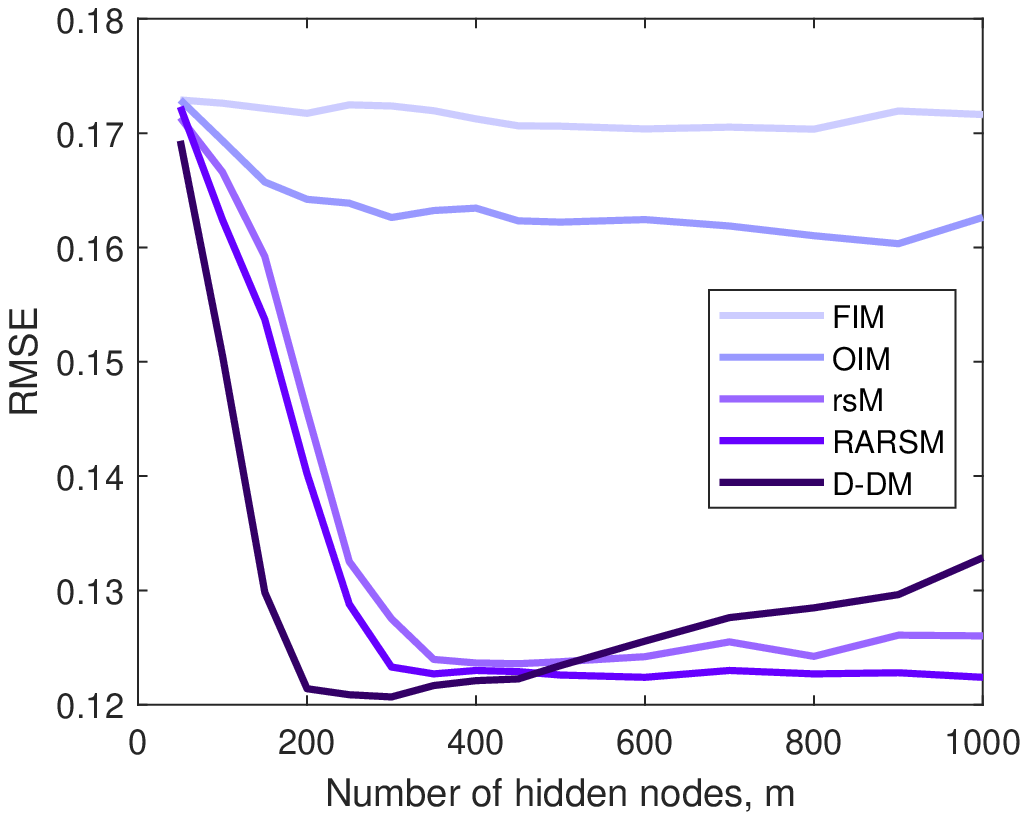}
	\includegraphics[width=0.49\textwidth]{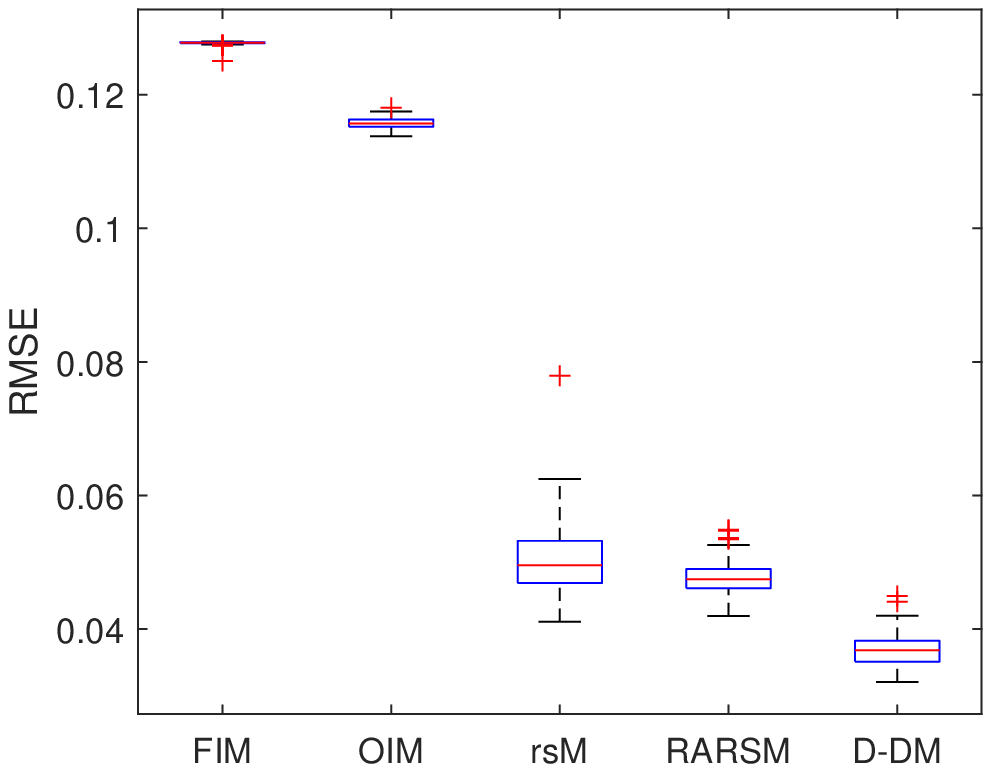}
	\caption{RMSE for different number of nodes $m$ (left panel) and distribution of the test RMSE for the tested methods (right panel).} 
	\label{fig7}
\end{figure}

First, we use function \eqref{eq25} with error level $c=0.2$ as the test function. The left panel of Fig. \ref{fig7} shows the cross-validation errors for different numbers of nodes at optimal values of other hyperparameters. As you can see in this figure, both FIM and OIM failed completely. Optimization of the interval bounds $[-u,u]$ in OIM brought only a slight improvement in accuracy when compared to FIM. More sophisticated methods, rsM, RARSM and D-DM, are incomparably more accurate. Note that the proposed D-DM needs the smallest number of neurons to get the best performance when compared to other methods. 

The right panel of Fig. \ref{fig7} shows the boxplots of the test RMSE for $100$ trials of the learning sessions carried out at the optimal values of hyperparameters. These simulations are summarized in Table 1. Clearly, from this table, D-DM outperforms the other methods in terms of accuracy. 
                     
In the next experiments we use two multivariable datasets: 
\begin{itemize}
	\item Stock – daily stock prices from January 1988 through October 1991, for ten aerospace companies. The task is to aproximate the price of the $10$-th company given the prices of the others (950 samples, 9 input variables, source: http://www.keel.es/).
	\item Kin8nm – a realistic simulation of the forward dynamics of an 8 link all-revolute robot arm. The task is to predict the distance of the end-effector from a target. The inputs are things like joint postions, twist angles, etc. (8192 samples, 8 input variables, source: www.cs.toronto.edu/~delve/data/kin /desc).  
\end{itemize}

\begin{figure}
	\centering
	\includegraphics[width=0.49\textwidth]{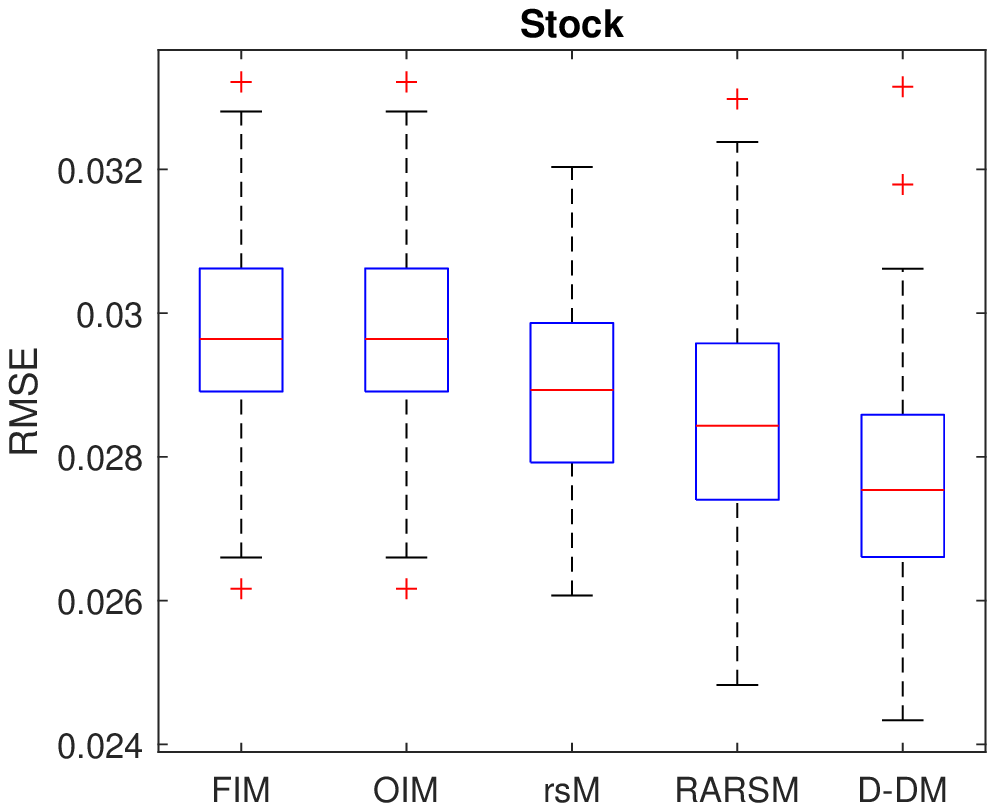}
	\includegraphics[width=0.49\textwidth]{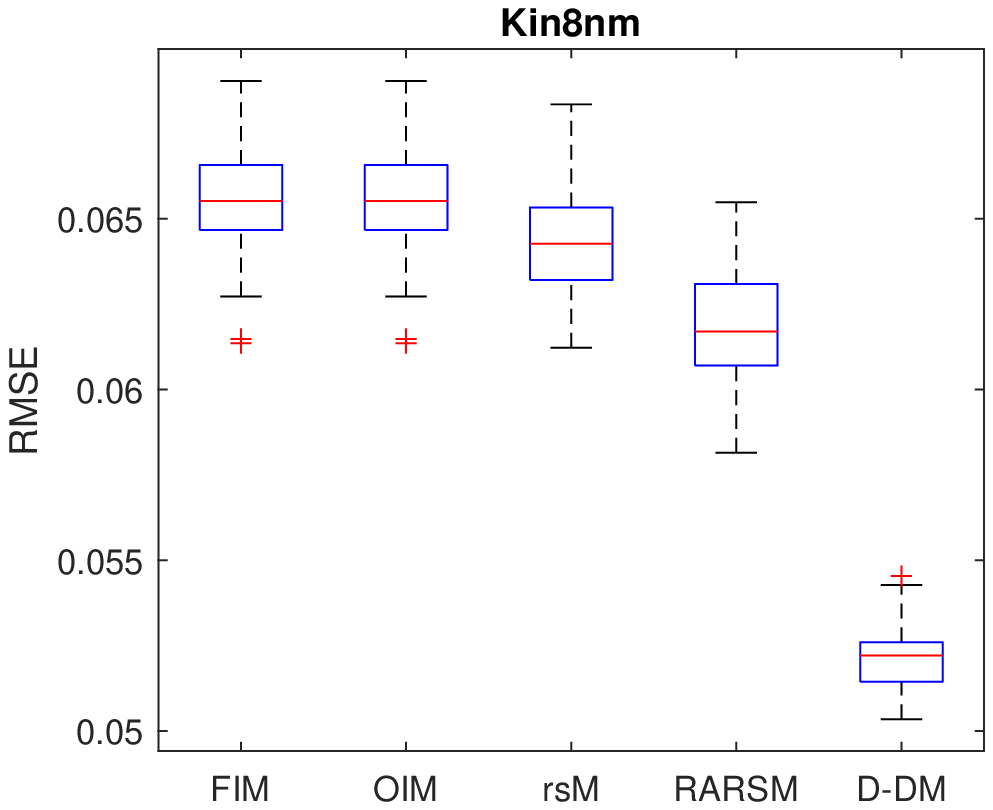}
	\caption{Distribution of the test RMSE for the Stock and Kin8nm data.} 
	\label{fig8}
\end{figure}

The data sets were divided into training sets containing $75\%$ samples selected randomly, and the test sets containing the remaining samples.
The test RMSE for both datasets at the optimal values of hyperparameters are visualized by the boxplots in Fig. \ref{fig8}. Table 1 shows the mean RMSE and the optimal hyperparameters of the methods. Note that D-DM demonstrates the best performance compared to other methods. Especially for Kin8nm the significant improvement in accuracy is observed for D-DM. This may be due to the fact that in this case the target function has variable fluctuations. The D-DM, which is designed to deal with such cases, performs best.   

\begin{table}
	\caption{Performance comparison of the proposed and comparative methods.}\label{tab1}
	\centering
	\begin{tabular}{|c|c|c|c|}
		\hline
		Method &  Test RMSE & \#nodes & Parameters\\
		\hline
		\textbf{Function \eqref{eq25}} & & & \\
		FIM &  $0.1277\pm 0.00029$ & 800 &-\\
		OIM &  $0.1157\pm 0.00088$ & 1000 & $u=3$\\
		rsM & $0.0503\pm 0.00527$ & 450 & $r=0.4, s=30$\\
		RARSM & $0.0477\pm 0.00254$ & 350 & $\alpha_{min}=55^\circ, \alpha_{max}=70^\circ$\\
		D-DM & $0.0370\pm 0.00230$ & 300 & $k'=35$\\
		\hline
		\textbf{Stock} & & & \\
		FIM &  $0.0296\pm 0.00141$ & 200 &-\\
		OIM &  $0.0296\pm 0.00141$ & 200 & $u=1$\\
		rsM & $0.0289\pm 0.00138$ & 200 & $r=0.001, s=2$\\
		RARSM & $0.0285\pm 0.00156$ & 200 & $\alpha_{min}=45^\circ, \alpha_{max}=70^\circ$\\
		D-DM & $0.0277\pm 0.00150$ & 250 & $k'=30$\\
		\hline
		\textbf{Kin8nm} & & & \\
		FIM &  $0.0655\pm 0.00153$ & 1300 &-\\
		OIM &  $0.0655\pm 0.00153$ & 1300 & $u=1$\\
		rsM & $0.0643\pm 0.00157$ & 1300 & $r=0.4, s=8$\\
		RARSM & $0.0618\pm 0.00157$ & 1300 & $\alpha_{min}=35^\circ, \alpha_{max}=55^\circ$\\
		D-DM & $0.0523\pm 0.00081$ & 900 & $k'=60$\\
		\hline
	\end{tabular}
\end{table}

\section{Conclusions}
The way in which the hidden node parameters are generated is a key issue in the randomized learning of FNN. When these parameters are selected in a standard way from the fixed interval the performance of the network can be weak, especially for complex function fitting.

This work proposes a new approach to generating the parameters of a FNN in randomized learning. The proposed method adjusts the hidden neurons weights and biases, representing the slopes and positions of the sigmoids, to the target function features. The method first randomly selects the input space regions by drawing the points from the training set. Then, the hyperplanes are fitted to the neighborhoods of the selected points and their coefficients are transformed into the sigmoid weights and biases. This results in the placement of the sigmoids in the selected regions of the input space and the adjustment of their slopes to the local fluctuations of the target function. As simulation research has shown, such a method of generating random parameters brings very good results in the approximation of the complex target functions when compared to the standard fixed interval method and other methods proposed 
recently in the literature.

Future work will focus on further analysis and improvement of the proposed method as well as rsM and RARSM, and their adaptation to classification problems.

%Przy rozszerzeniu zakresu mamy większe stromości funkcji i bardziej strome sigmoidy, które przemnożone przez wagi beta staja się jeszcze bardziej strome. Przy zawężeniu (poniżej przedziału [0,1]) funkcja się spłaszcza, kąty sa mniejsze. Po przemożeniu przez wagi beta sigmoidy się spłaszczają i nadmiernie wygładzają funkcję. 


\end{document}